\documentclass[10pt,leqno]{amsart}
\usepackage{graphicx}
\usepackage[none]{hyphenat}

\usepackage{indentfirst,csquotes}
\usepackage{hyperref}
\usepackage{caption}    
\usepackage{subcaption} 
\usepackage{placeins} 

\usepackage{amssymb,amsthm,amsmath}
\usepackage{xcolor,paralist,hyperref,titlesec,fancyhdr,etoolbox}

\titleformat{\section}[block]{\normalfont\Large\bfseries\centering}{\thesection}{1em}{}
\titleformat{\subsection}[runin]{\normalfont\normalsize\bfseries}{\thesubsection}{1em}{}
\titlespacing*{\subsection}{0pt}{3.25ex plus 1ex minus .2ex}{1em}

\hypersetup{ colorlinks=true, linkcolor=black, filecolor=black, urlcolor=black }

\usepackage{lipsum}

\begin{document}
\title{GameLabel-10K: Collecting Image Preference Data Through Mobile Game Crowdsourcing} 
\author[Initial Surname]{Jonathan Zhou}
\date{\today}
\email{jonathanzhou@aiapprenticeships.com, jonathantaozhou@gmail.com}
\maketitle

\begin{abstract}
  The rise of multi-billion parameter models has sparked an intense hunger for data across deep learning. This study explores the possibility of replacing paid annotators with video game players who are rewarded with in-game currency for good performance. We collaborate with the developers of a mobile historical strategy game, Armchair Commander, to test this idea. More specifically, the current study tests this idea using pairwise image preference data, typically used to fine-tune diffusion models. Using this method, we create GameLabel-10K, a dataset with slightly under 10 thousand labels and 7000 unique prompts. We fine-tune a model on this dataset, we fine-tune Flux Schnell and find an improvement in its prompt adherence, demonstrating the validity of our collection method. In addition, we publicly release both the dataset and our fine-tuned model on Huggingface. 
\end{abstract} 

\bigskip

\section{Introduction}
Deep learning models are notoriously data-hungry. Common approaches to gathering data include paying annotators to label data, collecting user preferences, and creating synthetic data \cite{1} \cite{2} \cite{3}. However, each of these methods has its  own limitations; paying annotators is often expensive, collecting user preferences requires a large group of users, and training on synthetic data increases the chance of mode collapse \cite{4}. In this paper, we explore another possibility: replacing certain advertisements with data labeling tasks. In particular, we focus on reward ads, a type of video game advertisement where players can watch an ad to receive in-game rewards. These advertisements typically have low conversion rates and profitability, reducing the opportunity cost of replacing them with data labeling activities. In particular, we focus on labeling prompt-image pairs, typically used for diffusion model fine tuning with algorithms such as Direct Preference Optimization (DPO) or Robust DPO \cite{5} \cite{6, 7, 8}. Each data point contains a prompt, two generated images, and, if the data point is annotated, at least one label indicating the preferred image. Our contributions are as follows. 

\begin{enumerate}
    \item We propose a novel method of data collection by replacing advertisements with data labeling tasks in mobile games.
    \item We implement this method in a fully real-world setting using the game Armchair Commander.
    \item We inspect the quality of data gathered and analyze the feasibility of our approach.
    \item We show that SOTA open-source diffusion models can be improved by crowdsourced data. 
    \item We release the dataset and model on Huggingface under a permissive Apache 2.0 License.\footnote{\url{https://huggingface.co/datasets/Jonathan-Zhou/GameLabel-10k}}
\end{enumerate}

\section{Related Work}
Data annotation is a well-known issue in the field of machine learning, and many attempts have been made to streamline this process. Previous attempts to crowdsource annotations, especially for image data, have found success, sometimes performing on par with expert annotators \cite{9}. 

Kirstain et al. crowdsourced a large image preference dataset from a group of users with great success \cite{10}. They gathered a large amount of Stable Diffusion users to interact with the model, creating a high-quality dataset. Riek et al. developed a Facebook game for crowd sourcing video annotations with promising results \cite{11}. They achieve high inter-rater reliability, with a Krippendorff’s $\alpha$ of 0.702, signalling high inter-rater reliability. However, their participants are volunteers recruited via word-of-mouth, which may perform differently from random players. 

Ponnada et al. have explored the idea of turning accelerometer data annotation into a videogame, where players annotate data as a recreational problem-solving exercise \cite{12}. The resulting data was of high-quality; however, without external rewards, players' incentives to annotate data strongly depend on their degree of engagement with the game. Additionally, the games designed for annotating activity recognition data were specifically built to entertain players; different types of data would require creating or altering games, which is time-consuming.

\section{Methodology}
\subsection{Data Type}
Data annotation is a broad field, ranging from video segmentation to language analysis. Our decision to work with preference data in the form of prompts and image pairs was made for multiple reasons. 
\begin{enumerate}
  \item Prompt-image pairs are relatively simple to obtain compared to other forms of data. 
  \item Preference labeling for images is a relatively low-skill task, so most players should be qualified annotators. 
  \item Humans tend to excel at image processing tasks, so data gathered might be of higher quality \cite{13}. 
  \item Diffusion models often have issues of poor prompt adherence, and we hope that our dataset will help diffusion model creators improve the prompt adherence of their models. 
\end{enumerate}

\subsection{Data Generation}
We first gathered relevant prompts from a dataset of 100,000 diffusion model prompts \cite{14}. To prevent inclusion of problematic content, we checked each prompt against a list of inappropriate words, removing all prompts that contained any elements from the list. This process reduced the number of available prompts from 100,000 to 30,000. Afterwards, we used Flux-Schnell \cite{15} to generate a pair of images for each prompt. 

\subsection{Comparison Data Points}
To preserve data integrity and prevent annotators from randomly labeling image pairs, a subset of our data contains data points with one image generated from the prompt and another image generated from an entirely different prompt. This makes the correct answer unambiguous for certain data points; we use these comparison data points to evaluate the quality of annotators. Accuracy is only calculated based on a labeler's performance comparison data points; most preference data is highly subjective and lacks a clear correct answer. Our goal is to filter out incompetent evaluators while allowing space for artistic and creative interpretation. 

\subsection{Data Collection}
We collaborated with the developer of a mobile strategy game, Armchair Commander, to recruit players and gather data. Our application is embedded within the game itself, requiring no additional downloads to run. Before labeling data, we show users a slideshow with instructions to select the image that best matches the prompt and informs them that their labels will be used in this study.

Once users have agreed to proceed, we send each user a single data point, which consists of a prompt and two images. The user labels the data point by selecting the higher quality image. Initially, we send comparison data points to assess the quality of data labeled by the user. If the user's accuracy is high, we slowly reduce the amount of comparison data. Annotators that perform poorly are permanently removed from the labeling platform and return to watching advertisements. Each user labels five data points at a time, and they are rewarded with in-game currency based on their accuracy.

Each annotator is presented with two images, and chooses between them without ties. We tried allowing annotators to select ties, but upon brief experimentation, we found that some annotators would default to selecting ties, even in cases when one image was clearly superior to another. To compensate for this discrepancy, we allow each data point to be labeled multiple times to obtain a better sample of the preference distribution. We also experimented with sending each annotator four images, but many players interact with Armchair Commander on their phone; users had trouble discerning details such as warped hands or limbs when four images were simultaneously displayed on a small screen.

For each user, we only store information that includes an anonymized user id, the amount of correctly and incorrectly labeled comparison data points, and which data points the user labelled. 

\section{Results}
\subsection{Quantity of Annotations}
In total, users annotated over 16,000 data points and around 6,000 were comparison images, leaving slightly under 10,000 labels in the dataset. The final dataset consists of approximately 6,800 data points, over 1,100 of which have been labeled multiple times. These images were collected over the course of three weeks. 

\subsection{User Statistics}
In total, 188 users were involved in this study, all of whom were random Armchair Commander players. The average user annotated slightly over 87 data points, with a standard deviation of approximately 194 data points. We found that a small subset of users performed most of the annotations; 16 users were responsible for nearly half of the annotations, with a single zealous user labeling over 2,000 images. Each annotator had an average accuracy of 74\%; however, since annotators that perform poorly are quickly eliminated, each data point is, on average, labelled by an annotator with an accuracy of 83\%.

\subsection{Estimated Cost} The estimated opportunity cost of this experiment was approximately \$8 USD; this number was calculated based on the game's mean revenue per reward advertisement. We do not include potential sources of indirect revenue losses, such as in-game currency losing value, because Armchair Commander's revenue greatly fluctuates across different months; attempts to measure these losses did not yield statistically significant results. 

\subsection{Visual Examples}
We found that annotators often struggle with prompts that mention specific nouns, particularly when they reference obscure celebrities, historical figures, or unfamiliar art styles. Despite these challenges, we believe the overall quality of the dataset remains high; most instances of physically impossible behavior, such as people inside of walls, have been identified. Additionally, the majority of instances of images that do not adhere to the prompt have been properly labeled. Below are some random (not cherry picked) examples of our data.

\begin{figure}[htbp]
  \centering
  \begin{minipage}{1\textwidth} 
      \centering
      \text{Prompt: Another planet alongside an aqualung in the style of a movie poster}\\[0.1cm]
      \begin{subfigure}[b]{0.25\textwidth}
          \centering
          \includegraphics[width=\textwidth]{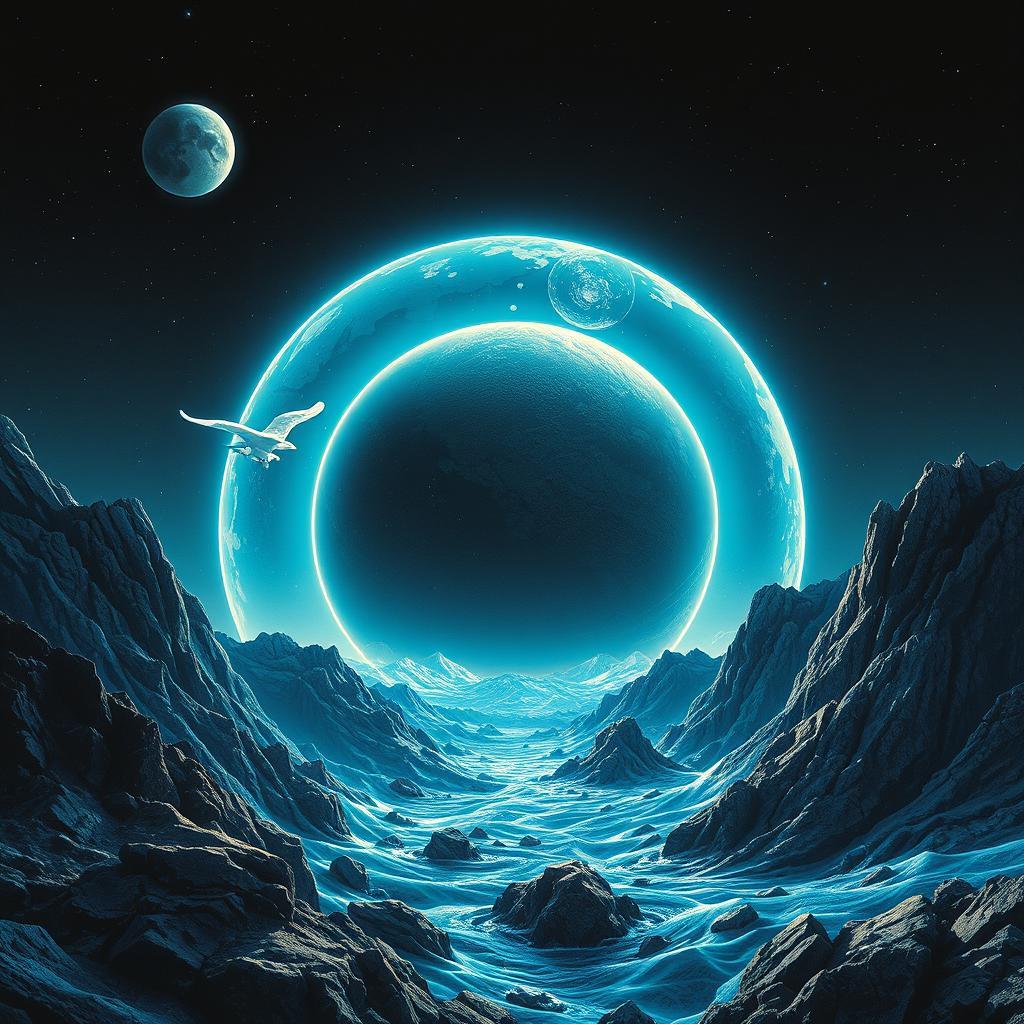}
          \caption*{Preferred}
      \end{subfigure}
      \hspace{0.02\textwidth} 
      \begin{subfigure}[b]{0.25\textwidth}
          \centering
          \includegraphics[width=\textwidth]{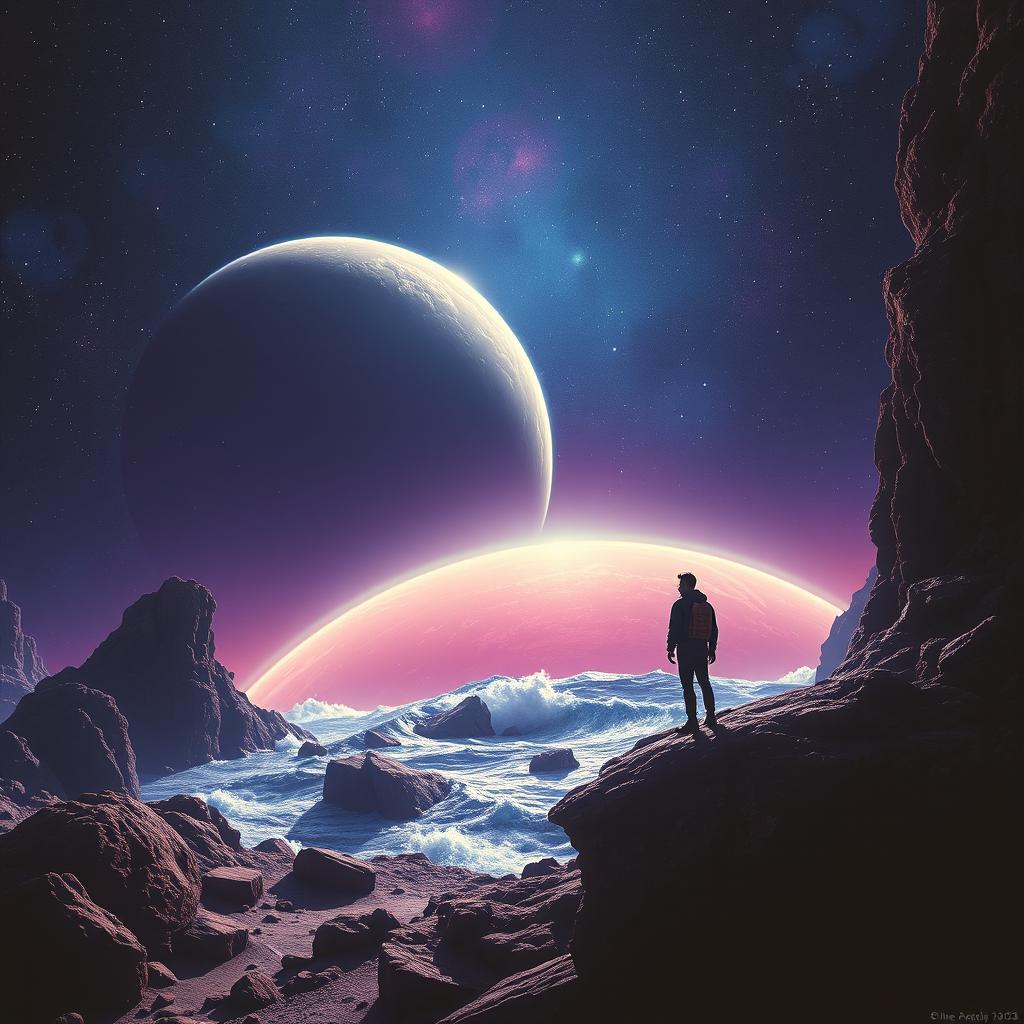}
          \caption*{Dispreferred}
      \end{subfigure}
  \end{minipage}

  \vspace{1cm} 

  \begin{minipage}{1\textwidth}
      \centering
      \text{Prompt: Alien sitting on top of a rug in the style of a crayon drawing}\\[0.5cm]
      \begin{subfigure}[b]{0.25\textwidth}
          \centering
          \includegraphics[width=\textwidth]{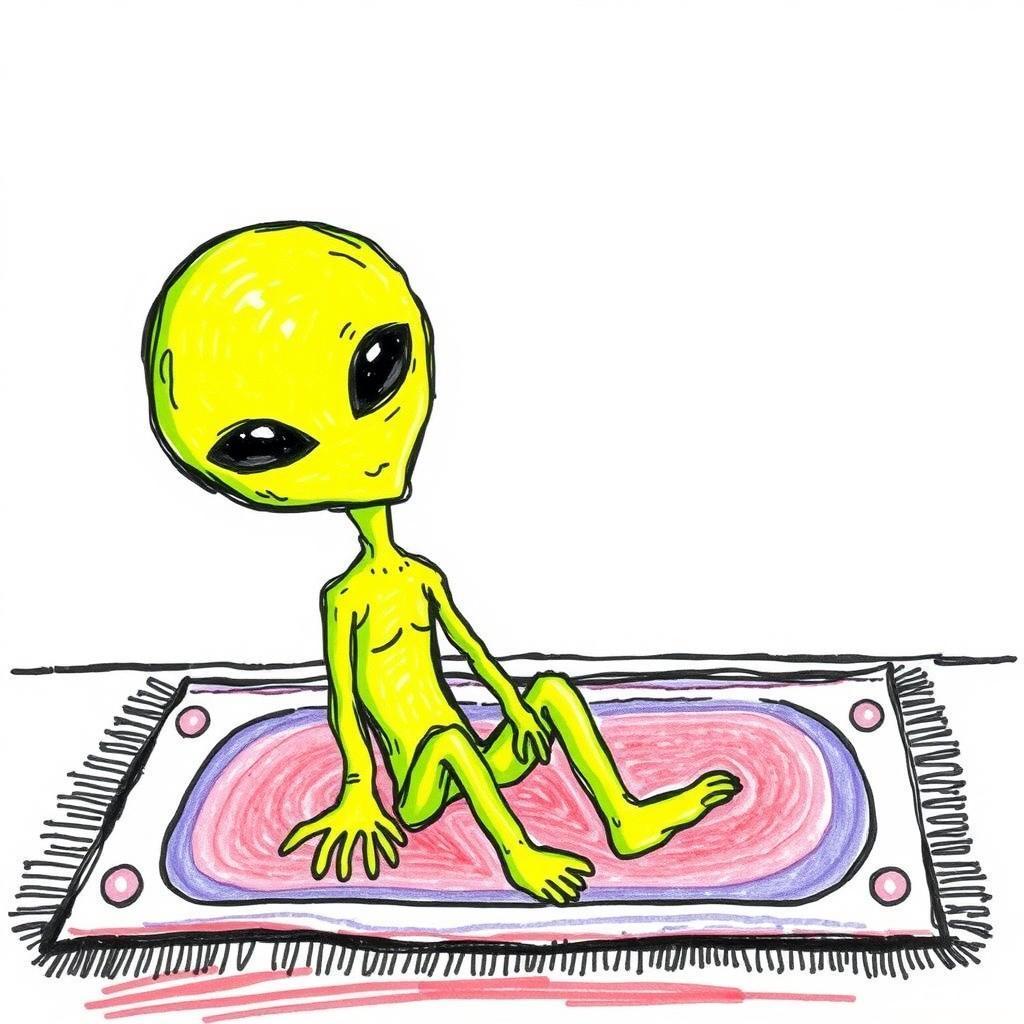}
          \caption*{Dispreferred}
      \end{subfigure}
      \hspace{0.02\textwidth} 
      \begin{subfigure}[b]{0.25\textwidth}
          \centering
          \includegraphics[width=\textwidth]{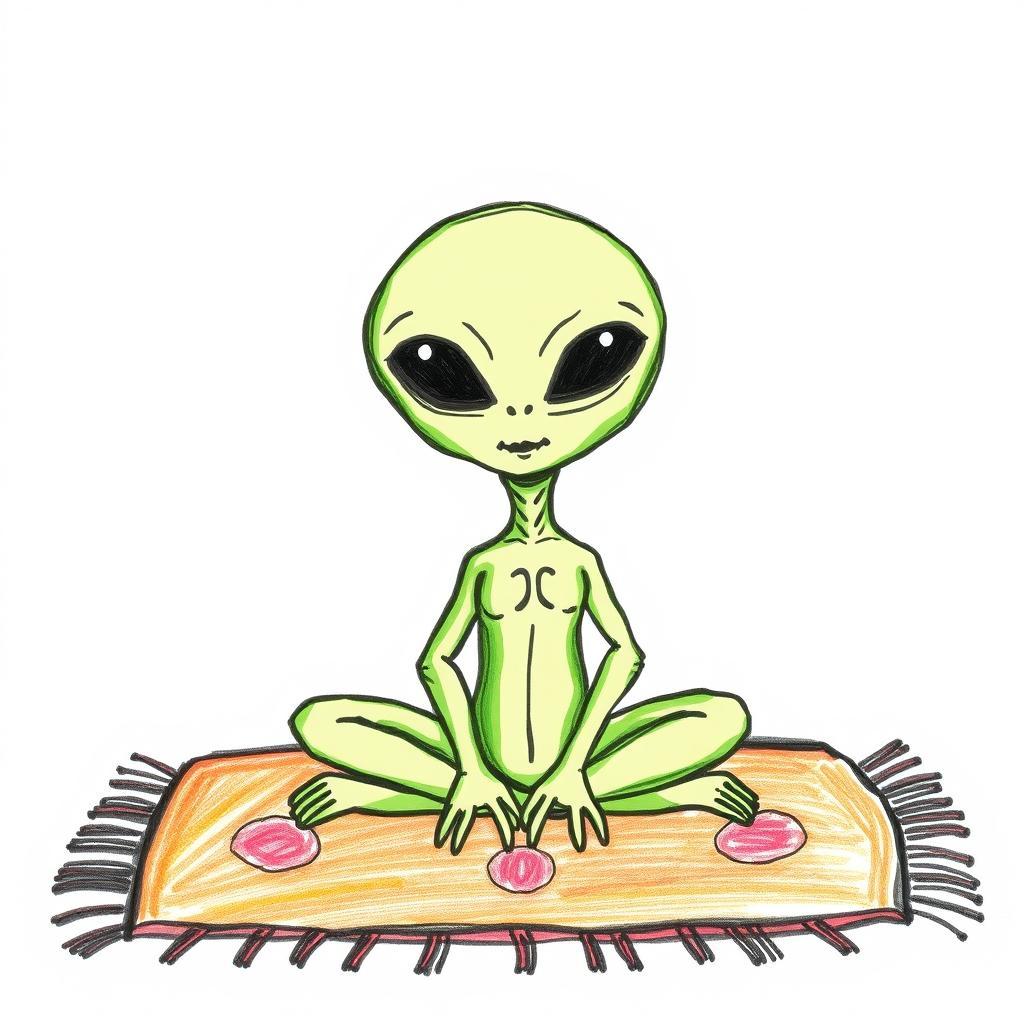}
          \caption*{Preferred}
      \end{subfigure}
  \end{minipage}
  \vspace{1cm}

  \begin{minipage}{1\textwidth}
    \centering
    \text{Prompt: Cave drawing of Historian and an eagle sitting by some trees}\\[0.5cm]
    \begin{subfigure}[b]{0.25\textwidth}
        \centering
        \includegraphics[width=\textwidth]{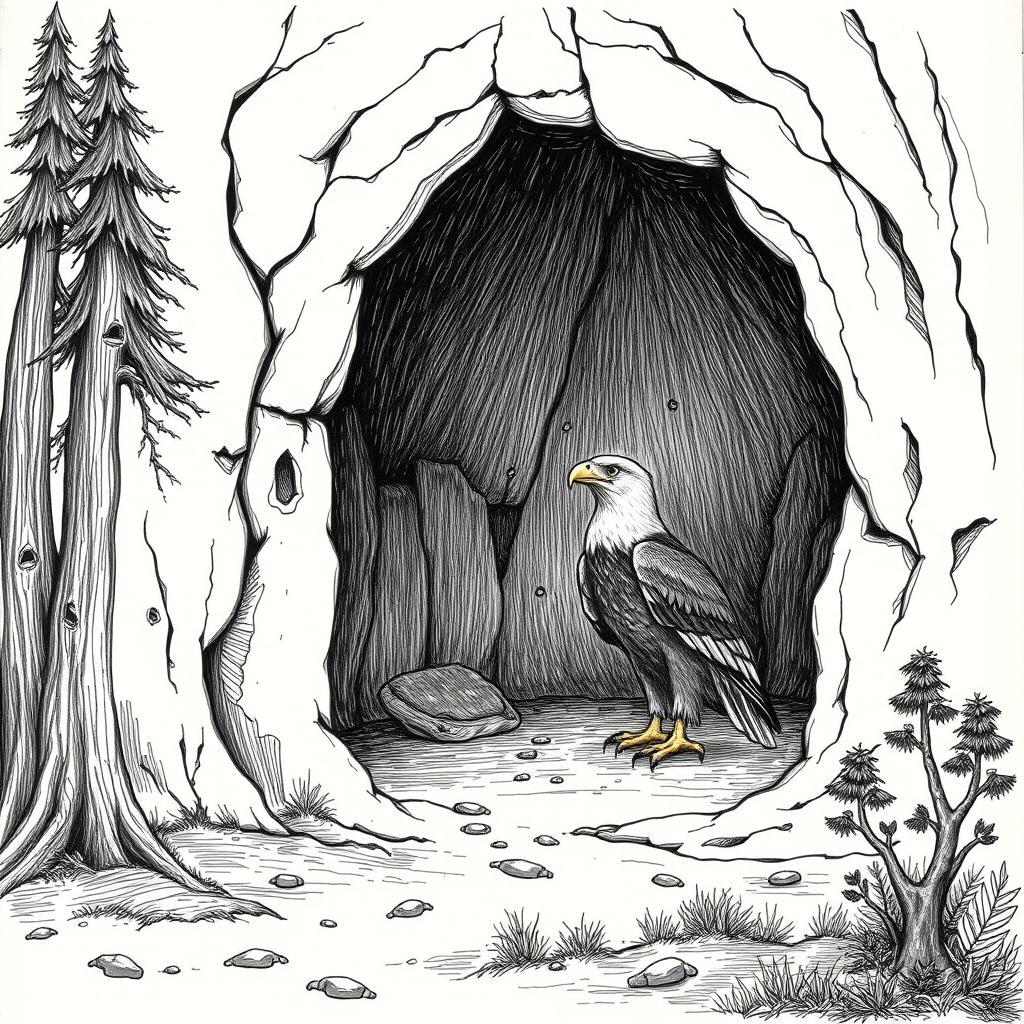}
        \caption*{Dispreferred}
    \end{subfigure}
    \hspace{0.02\textwidth} 
    \begin{subfigure}[b]{0.25\textwidth}
        \centering
        \includegraphics[width=\textwidth]{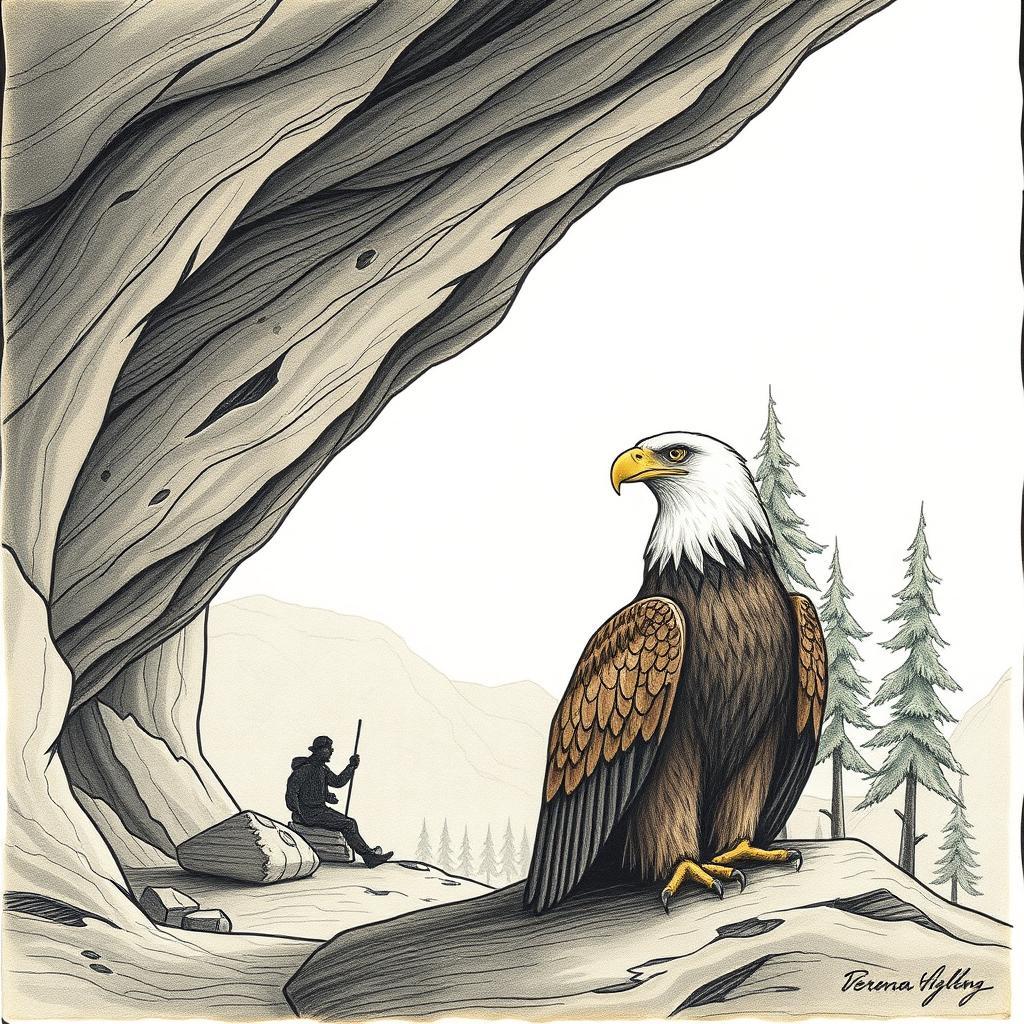}
        \caption*{Preferred}
    \end{subfigure}
\end{minipage}

\end{figure}

\FloatBarrier

\subsection{Finetuning Flux on Examples}
In addition to manually reviewing the results of this dataset, we fine-tune Flux-Schnell on data GameLabel-10K to empirically demonstrate the validity of our collection method. More specifically, we fine-tune Flux with LORA \cite{16} on the subset of preferred images from our dataset, which we find improves the model's ability to follow text based instructions. We open source the resulting LORA on Huggingface, along with the code for both training and inference.\footnote{\url{https://huggingface.co/Jonathan-Zhou/Flux-GameLabel-Lora}}  However, due to retraining on generated images, we find notable degradations in image quality unrelated to the quality of our labels; we verify this by fine-tuning Flux on its own images and observe a similar degree of image quality loss. We hypothesize that more refined methods, such as RLHF, would result in higher quality models. Below are some comparisons between our fine-tune and the original model. Images on the left are from our fine-tune, and images on the right are from the original Flux-Schnell. 

\begin{figure}[htbp]
    \begin{minipage}{1\textwidth}
        \centering
        \begin{subfigure}[b]{0.7\textwidth}
            \centering
            \includegraphics[width=\textwidth]{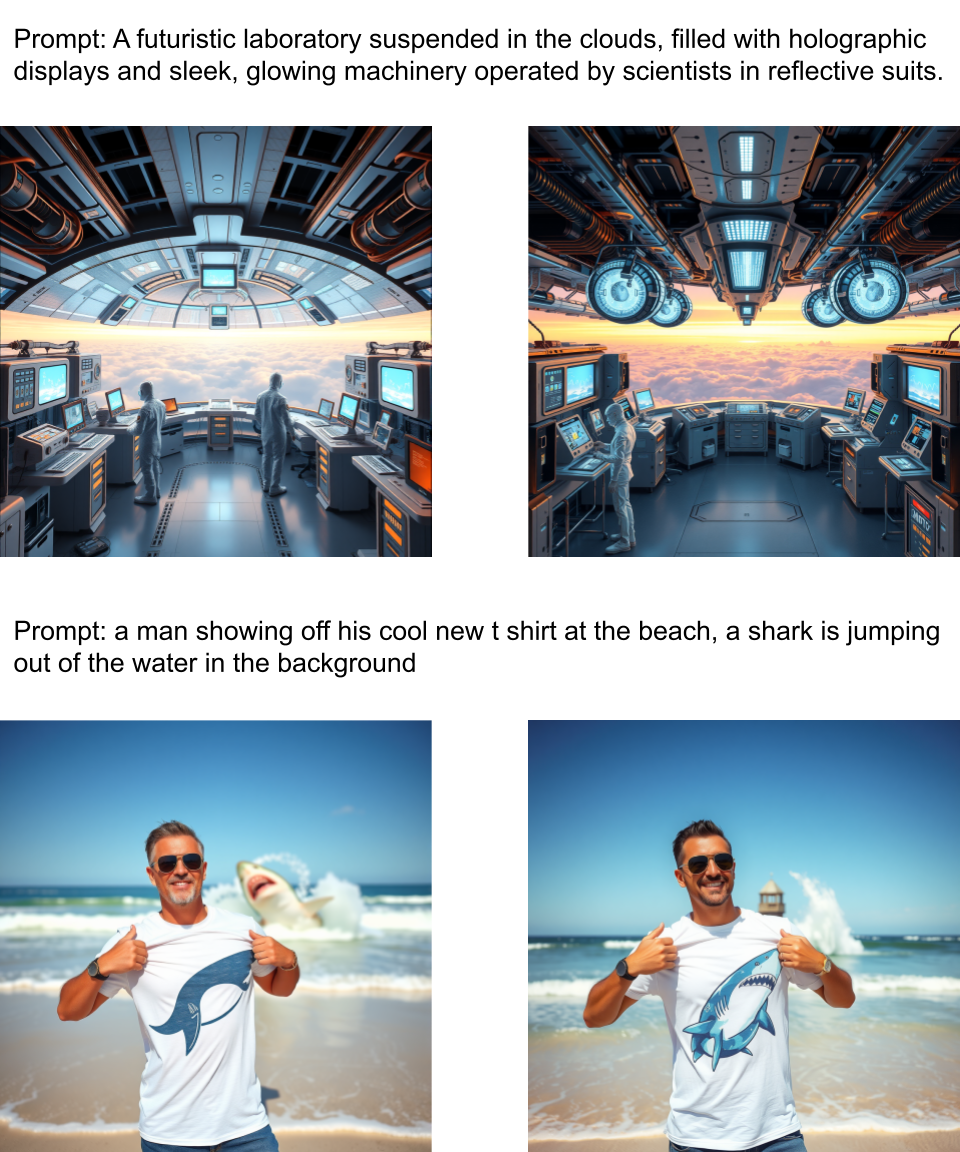}
        \end{subfigure}
        \hspace{0.02\textwidth} 
    \end{minipage}
    
    \end{figure}
    
    \FloatBarrier

\section{Future Work}
We believe much work remains in this area. Adding a timer that prevents users from submitting annotations stands out as a straightforward method of improving data quality. We hypothesize that many users do not read the prompt in detail, causing them to select the wrong image. A timer could encourage users to prioritize quality over quantity and lead to significant improvements in data quality. An interactive tutorial may be another way to improve data quality; the instructions we send users are dry, and an interactive tutorial is likely to improve retention. Capturing more detailed user statistics, such as recording the amount of time users spend annotating each data point, would likely be helpful in minimizing the cost of gathering data. Additionally, contacting and surveying users could measure the impact labeling data has on the overall game experience and provide insight on possible improvements to this system.

Another research direction is analyzing the relationship between the game used to collect data and the quality of data collected. User demographics vary across games, and players from different games may have differences that affect their annotations. Armchair Commander is a historical war strategy game, and we hypothesize that players of historical games likely possess superior historical knowledge compared to the general population of video game players, leading to more accurate annotations. A study that implements this framework across multiple games could provide valuable insight on this topic, which could be later used to send data to the most suitable playerbase and optimize the process of data collection.

\end{document}